\documentclass[conference]{IEEEtran}
\usepackage{cite}
\usepackage{amsmath,amssymb,amsfonts}
\usepackage{algorithmic}
\usepackage{graphicx}
\usepackage{textcomp}
\usepackage{xcolor}
\def\BibTeX{{\rm B\kern-.05em{\sc i\kern-.025em b}\kern-.08em
    T\kern-.1667em\lower.7ex\hbox{E}\kern-.125emX}}
\begin{document}

\title{A Semi-Supervised Generative Adversarial Network for Prediction of Genetic Disease Outcomes}

\author{
\IEEEauthorblockN{Caio Davi}
\IEEEauthorblockA{\textit{Department of Electrical and Computer
Engineering} \\
\textit{Texas A\&M University}\\
College Station, United States \\
caio.davi@tamu.edu
 }

\and

\IEEEauthorblockN{ Ulisses Braga-Neto}
\IEEEauthorblockA{\textit{Department of Electrical and Computer
Engineering} \\
\textit{Texas A\&M University}\\
College Station, United States \\
ulisses@tamu.edu
}
}

\maketitle

\begin{abstract}
For most diseases, building large databases of labeled genetic data is an expensive and time-demanding task. To address this, we introduce genetic Generative Adversarial Networks (gGAN), a semi-supervised approach based on an innovative GAN architecture to create large synthetic genetic data sets starting with a small amount of labeled data and a large amount of unlabeled data. Our goal is to determine the propensity of a new individual to develop the severe form of the illness from their genetic profile alone. 
The proposed model achieved satisfactory results using real genetic data from different datasets and populations, in which the test populations may not have the same genetic profiles. The proposed model is self-aware and capable of determining whether a new genetic profile has enough compatibility with the data on which the network was trained and is thus suitable for prediction. The code and datasets used can be found at (suppressed in anonymous version)
{\tt https://github.com/caio-davi/gGAN}
\end{abstract}


\section{Introduction}

Creating large genetic datasets is an expensive and time-demanding task in the case of most diseases, as it involves patient recruitment, laboratory work, sequencing equipment, and bioinformatics expertise. In many instances, data is classified as Protected Health Information (PHI)\cite{assistance2003summary}, making it difficult to obtain and to handle. There is also a range of intrinsic characteristics within this domain that increase the complexity of archiving and handling this data. Firstly, it may be difficult to obtain patients from the right population to extract the genetic material from. Secondly, maintaining the quality of the analysis is not a trivial process due to biological and technical noise, which varies between laboratories and/or between batches.

With current advancements in Machine Learning algorithms applied to human genetics \cite{sheehan2016deep,schrider2018supervised,flagel2018unreasonable}, large datasets are increasingly needed in order to support the use and analysis of a variety of data-intensive algorithms. Therefore, new mechanisms that are capable of generating coherent synthetic genetic data would be a remarkable advancement within this field.

However, genetic data on neglected diseases is notoriously scarce. This is the case of Dengue disease, a viral disease most commonly found in tropical areas of under-developed and developing countries. 
Dengue is a global public health concern with an annual global incidence of 390 million cases\cite{bhatt2013global}. There is evidence supporting that the disease progression is genetically influenced\cite{xavier2017host,useche2019association,just1995genetic,czaja1993genetic}, but, as mentioned previously,  genetic data on Dengue is scarce. 

Recently, we employed machine learning algorithms to find associations between profiles of single-nucleotide polymorphisms (SNPs) and disease severity using a small data set from a cohort of patients in Recife, Brazil \cite{davi2019severe}. In this paper, we propose gGAN, an innovative Generative Adversarial Network (GAN) architecture to create large synthetic genetic data starting with the small amount of labeled data used in the previous study and a large amount of unlabeled data from the human 1000 genome project. This leads to a semi-supervised classifier that is able to predict the genetic predisposition to severe dengue of a patient, even if their genetic profile is from a different population than the one originally used to train the classifier (in this case, a Brazilian population). The classifier is self-aware in that one of the outputs of the proposed GAN is able to discriminate whether a given test genetic profile is sufficiently compatible with the genetic profiles in the training population. Even though we use dengue to demonstrate our method, it is entirely general and can be applied to other diseases.

\begin{figure*}[t!]
    \begin{center}
        \includegraphics[width=0.8\textwidth]{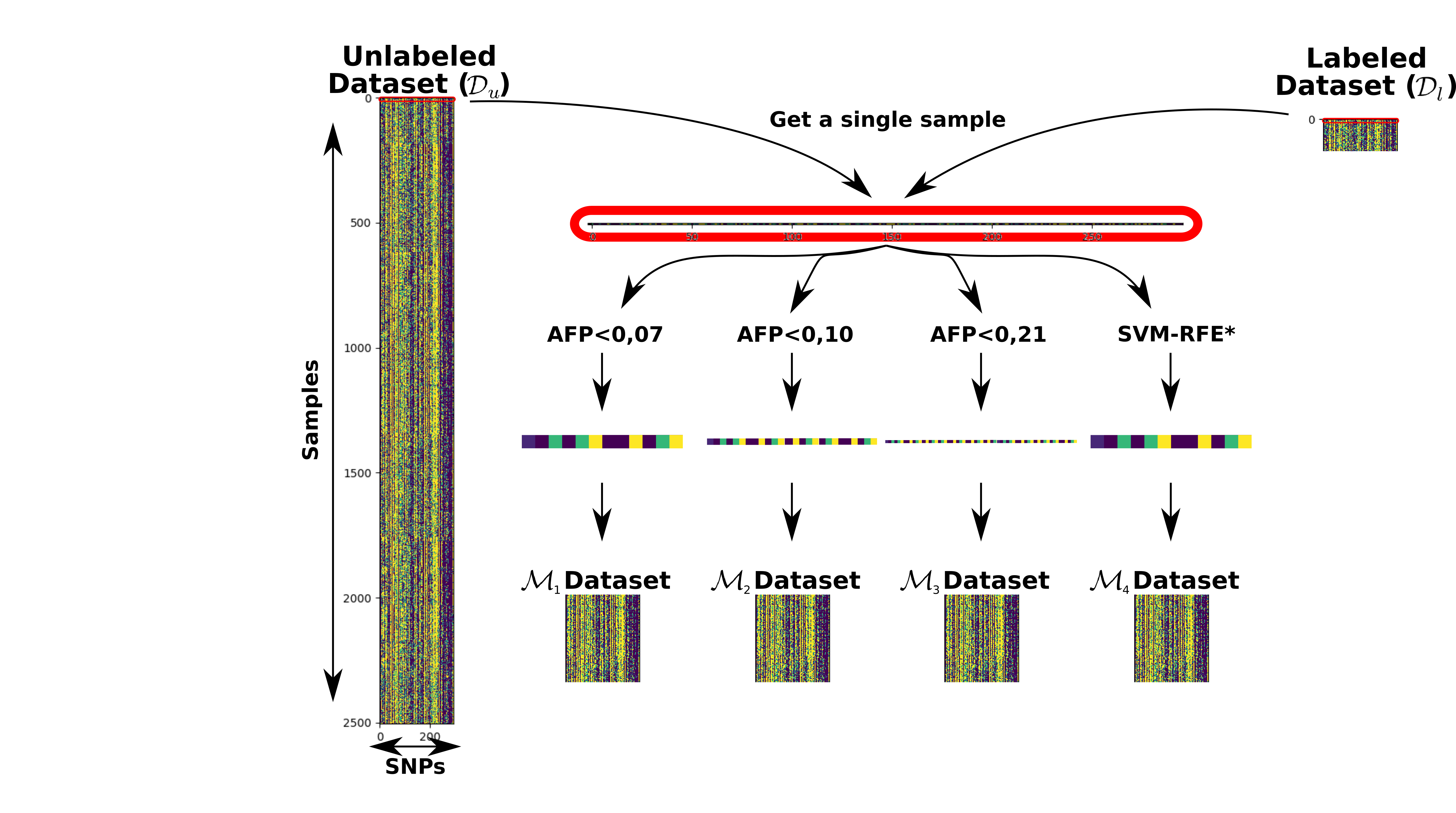}
   \end{center}
\vspace{-3ex}
   \caption{Data preprocessing procedure. Each line in each dataset represents a genetic profile of a patient. The SNPs are selected according to an Allelic Frequency Distance (AFD) threshold and the SVM-RFE algorithm in \cite{davi2019severe}. In this figure, AFD thresholds of 0.07, 0.10, 0.21 are depicted.}
\label{fig:sampling}
\end{figure*}

\section{Background}

Genome-wide association studies (GWAS) are a powerful tool for identifying genetic influences in diseases, particularly for conditions influenced by a variety of genes and environmental factors at once \cite{manolio2009cohort}.
These techniques are historically used for finding genetic differences in case-controlled studies for particular diseases.

One of the biggest concerns related to genetic studies is the size of the patient cohort used in the experiments. As mentioned previously, obtaining a patient cohort is a hard and expensive task. Moreover, the discrepancy between the size of the cohort and the number of features is a frequent burden. This occurs due to the large number of Single Nucleotide Polymorphisms (SNPs) in the human genome, which, on average, occur almost once in every 1,000 nucleotides. This means there are roughly 4 to 5 million SNPs in a person's genome. This factor can easily lead to overfitting due to the curse of dimensionality \cite{keogh2010curse}. To overcome this, feature selection becomes necessary.

Dengue disease is a global public health concern caused by the Dengue Virus (DENV), a positive-sense RNA virus from the \textit{Flaviviridae} family.
According to the 2009 World Health Organization (WHO) Dengue classification guideline\cite{world2009dengue}, symptomatic infection by DENV can range from a mild disease know as Dengue Fever (DF), to a more severe disease known as Severe Dengue (SD), which displays hemorrhagic complications that may eventually culminate in the death of the host individual.
 
There is evidence that some individuals may be genetically predisposed to some prognostic of the Dengue disease \cite{xavier2017host,useche2019association}. More specifically, studies have found associations between the genetic expression, mostly in the innate and in the adaptive response pathways, and the Dengue infection phenotype \cite{nascimento2009gene,brasier2012three,brasier2015molecular,chatterjee2018clinical,azevedo2019aa}. Also, there are efforts to use Machine Learning techniques to find associations between static genetic profiles and dengue severity \cite{gomes2010classification,davi2019severe}.

Generative Adversarial Networks (GAN) \cite{goodfellow2014generative} are a promising ML approach to generate synthetic data. GANs are commonly used to generate synthetic images from a known training set. However, there are applications of GANs in other domains\cite{esteban2017real}, including genetics\cite{yelmen2019creating}. There is also a report of a semi-supervised approach to GANs \cite{salimans2016improved}, for classification of the MNIST dataset \cite{lecun1998gradient}.

\section{Methods}\label{methods}

This section describes the steps performed  to create the model. We begin with pre-processing of our datasets, then describe the model itself, and finally how it was trained. The tests and results are described in the next section. 

\subsection{Data Preprocessing}\label{preprocessing}

We used two primary datasets from real genetic data available to the research community. The first is a dataset labeled with the Dengue Infection phenotype in human subjects. Patients with dengue-related symptoms were screened from three hospitals in the city of Recife, Brazil; this population constitutes the domain $\mathcal{D}_l$. The first dataset, denoted as $\mathcal{X}_l$, was extracted from this population, and it consists of 102 patient genotypes measured at 322 loci polymorphisms. The methods of extraction and genotyping and cohort description for the labeled dataset can be found in \cite{cordeiro2007characterization}.

\begin{figure*}[t!]
    \begin{center}
        \includegraphics[width=0.8\textwidth]{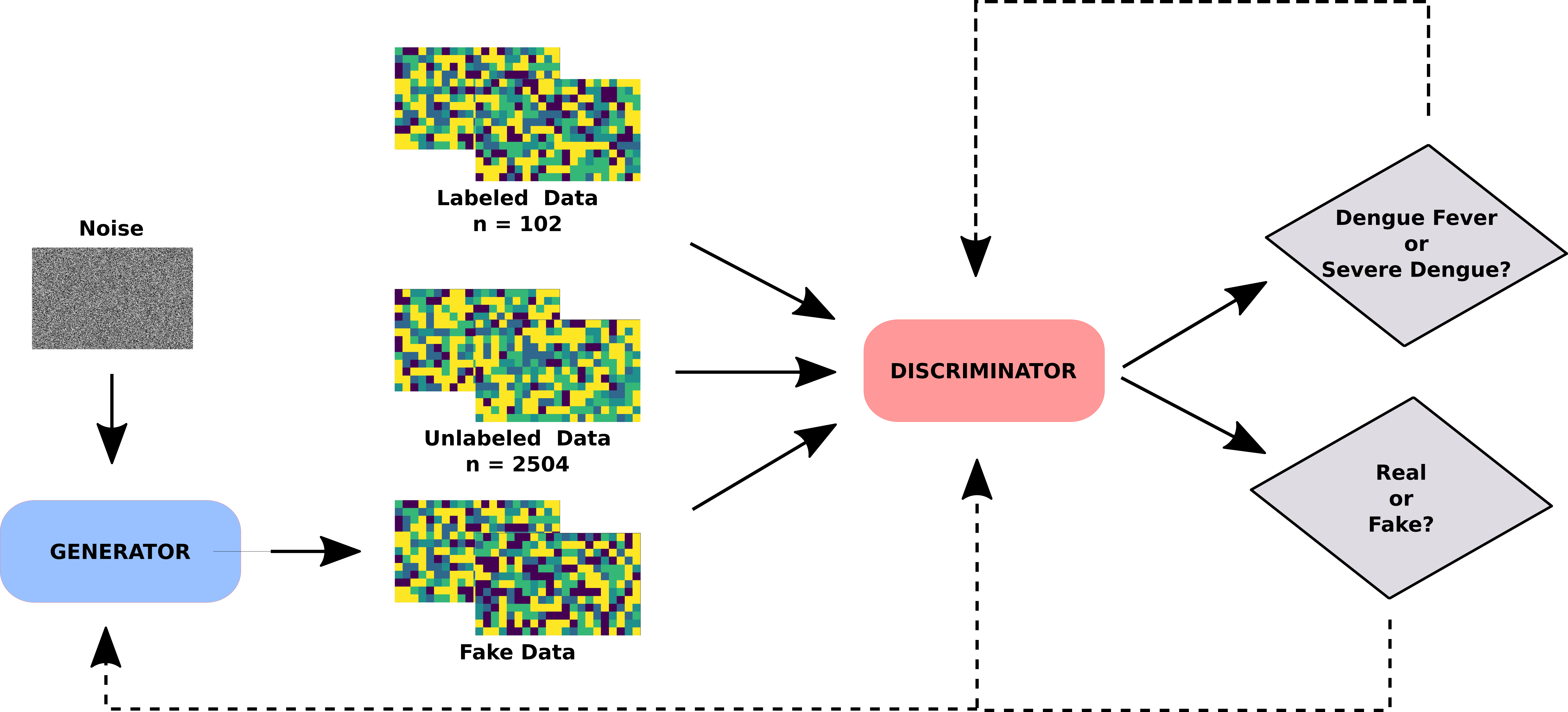}
   \end{center}
   \caption{Model architecture. The discriminator network is fed two datasets (labeled dataset (n=102), unlabeled dataset (n=2504)) and the synthetic genetic profiles from the generator network. The discriminator training consists of two phases, the first using real unlabeled data and synthetic data, and the second using labeled data.}
\label{fig:model}
\end{figure*}

The second dataset contains unlabeled genetic data from Phase 3 of the 1000 Genomes Project\cite{10002015global} (which consists of 2504 individuals genotyped for more than 84 million variants). The 1000 Genomes Project maps genetic information from populations around the world; we call this population $\mathcal{D}_u$. The unlabeled dataset $\mathcal{X}_u$ consists of the same set of 322 loci in the labeled dataset, extracted from the 1000 Genomes Collection. An important step in the manipulation of this kind of data is that we estimated each of the allelic frequencies\cite{richards2015standards}. Different populations tend to have different frequencies due to their diverse genomic background\cite{naslavsky2017exomic}. These differences may affect the performance of the discriminator network, which would try to learn from two datasets with different frequency distributions.  With this in mind, we compared these frequencies between the labeled dataset and the unlabeled dataset. We measured the frequencies for each allele of each variant in both datasets and assume the largest difference between them as their frequency proximity. The Allelic Frequency Distance (AFD) between two domains is defined as
\begin{equation}
{\rm AFD} \,\,= \max({f_i}(\mathcal{X}_l)-{f_i}(\mathcal{X}_u))\,.
\label{eq:AFP}
\end{equation}
In equation (1), $f_i$ is the frequency of a specific allele $i$ within the $n_i$ possible values for each nucleotide in population $\mathcal{X}$.

After this, we established certain thresholds to sampling the data and used them to create different INPUT configurations to the network. We used thresholds of 0.7, 0.1, and 0.21 resulting in subsets of 12, 25, and 96 SNPs, as shown in Figure \ref{fig:sampling}. Those subsets were used to build the models $\mathcal{M}_1$, $\mathcal{M}_2$ and $\mathcal{M}_3$, respectively.

We also used a subset of SNPs chosen in \cite{davi2019severe}. In that paper, a Support Vector Machine was used to find the best SNPs to discriminate Severe Dengue. As a result, 13 SNPs were found as the most adequate ones to be used to classify the severe form of the disease. One of those SNPs was not genotyped in the 1000 Genome Project, so it was removed from this set, resulting in 12 SNPs for model $\mathcal{M}_4$.

\subsection{gGAN Architecture}\label{archi}

We modified the traditional GAN architecture to work with both labeled and unlabeled data. In the traditional GAN model, the discriminator has only one output, which classifies the input as real or fake (synthetic). 
In our case, we want to predict the label of a test genetic profile ("Dengue Fever" or "Severe Dengue") in addition to classifying between real or fake. In order to archive this, we added a new output to the discriminator network. This is illustrated in Figure \ref{fig:model}. Tables~\ref{tab:disc_model}~and~\ref{tab:gen_model}  give a detailed overview of the layers of the discriminator and generator networks in the implementation of the proposed paradigm. All layers and operators in these tables are described in Tensorflow API notation. Notice that all generators have inputs of the same length. The generators are always fed with random noise, and in this case, we opted to keep the same size for all the models. The Upsample layer simply increases the size of the layer by duplicating it.

\begin{table}[h]
\caption{Architecture of the discriminator networks.}
\label{tab:disc_model} 
\begin{tabular}{ccccc}
\hline
 $\mathcal{M}_{1,disc}$&  $\mathcal{M}_{2,disc}$ &  $\mathcal{M}_{3,disc}$ &  $\mathcal{M}_{4,disc}$ \\ \hline
input(12,1)    & input(25,1)     &  input(96,1)  &  input(12,1)   \\
ReLu(12,24)    & ReLu(25,50)     &  ReLu(96,180) &  ReLu(12,24)   \\
ReLu(12,48)    & ReLu(25,100)    &  ReLu(96,360) &  ReLu(12,48)   \\
ReLu(12,24)    & ReLu(25,50)     &  ReLu(96,180) &  ReLu(12,24)   \\
Flatten        & Flatten         &  Flatten      &  Flatten       \\
Dropout        & Dropout         &  Dropout      &  Dropout       \\
Output$_{l}(2,1)$& Output$_{l}(2,1)$&  Output$_{l}(2,1)$&  Output$_{l}(2,1)$  \\
Output$_{u}(2,1)$& Output$_{u}(2,1)$&  Output$_{u}(2,1)$&  Output$_{u}(2,1)$  \\\hline
\end{tabular}
\end{table}

\begin{table}[h]
\caption{Architecture of the generator networks.}
\label{tab:gen_model} 
\begin{tabular}{ccccc}
\hline
 $\mathcal{M}_{1,gen}$&  $\mathcal{M}_{2,gen}$ &  $\mathcal{M}_{3,gen}$ &  $\mathcal{M}_{4,gen}$ \\ \hline
input(100)    &  input(100)     &  input(100)     &input(100)    \\
ReLU(24)      &  ReLU(50)       &  ReLU(90)       &ReLU(24)      \\
UpSample(48)  &  UpSample(100)  &  UpSample(180)  &UpSample(48)  \\
ReLU(48)      &  ReLU(100)      &  ReLU(180)      & ReLU(48)    \\ 
UpSample(96)  &  UpSample(200)  &  UpSample(360)  &UpSample(96)  \\
ReLu(12)      &  ReLu(25)       &  ReLu(96)       &ReLu(12)      \\
Reshape(12,1) &  Reshape(25,1)  &  Reshape(96,1)  &Reshape(12,1) \\ \hline
\end{tabular}
\end{table}

During training, the discriminator network weights are updated using the labeled and unlabeled data. This update occurs in turns (labeled and then unlabeled) for each epoch of training. First the network $\mathcal{M}_{m,disc}$, for $m=1,\ldots,4$, is exposed to the labeled samples in $\mathcal{D}_l$, and a loss $\mathcal{L}(\mathcal{D}_l)$ is calculated using the binary cross-entropy \cite{zhang2018generalized}:
\begin{equation}
\mathcal{L}(\mathcal{D}_l) \,=\, - \frac{1}{N} \sum_{i=1}^N \sum_{j=0}^{1} y_{ij}\log\hat{y}_{ij}\,,
\label{eq:LabeledLoss}
\end{equation}
where $N$ is the batch size, $(y_{i0},y_{i1})$ is a one-hot encoding for the training label ("Dengue Fever" or "Severe Dengue"), and $(\hat{y}_{i0},\hat{y}_{i1})$ is the softmax output of the discriminator corresponding to training profile ${\mathbf x}_i$, for $i=1,\ldots,N$. 
Next, the weights are updated using profiles in $\mathcal{D}_u$ using the loss function:
\begin{equation}
\mathcal{L}(\mathcal{D}_u, \mathcal{D}_g) \,=\, - \frac{1}{M} \sum_{i=1}^M \sum_{j=0}^{1} y_{ij}\log\hat{y}_{ij}
\label{eq:LabeledLoss}
\end{equation}
where $M$ is the batch size, $(y_{i0},y_{i1})$ is a one-hot encoding for the training label (``fake'' or ``real''), and $(\hat{y}_{i0},\hat{y}_{i1})$ is the softmax output of the discriminator corresponding to training profile ${\mathbf x}_i$, for $i=1,\ldots,M$. The training batch is composed by unlabeled real samples ($\mathcal{D}_u$) and synthetic samples ($\mathcal{D}_g$) created by the generator model. 

On the other hand, the weights of the generator network are updated using random noise as the input data set, denoted by $\mathcal{D}_r$. A synthetic sample will be produced using this random input and then the discriminator will classify it as real or fake data. Based on this answer, the binary cross-entropy is used to calculate the loss:
\begin{equation}
\mathcal{L}(\mathcal{D}_r) = - \frac{1}{P} \sum_{i=1}^P  \sum_{j=0}^{1} y_{ij}\log\hat{y}_{ij}
\label{eq:genLoss}
\end{equation}
where $P$ is the batch size, $(y_{i0},y_{i1}) = (0,1)$ is a one-hot encoding for the training label (all profiles are ``fake''), and $(\hat{y}_{i0},\hat{y}_{i1})$ is the softmax output of the discriminator corresponding to the random genetic profile ${\mathbf x}_i$, for $i=1,\ldots,P$. 

\section{Results}

The dataset from the labeled domain $\mathcal{D}_l$ was split into training and testing following a proportion of 80\% for training and the remaining 20\% for testing. This resulted in a training dataset $\mathcal{X}_{l_{train}}$ containing 72 labeled genetic profiles, and testing dataset $\mathcal{X}_{l_{test}}$ with the remaining 20 labeled profiles. We used the whole unlabeled dataset to train during the unlabeled step and selected 20\% as a testing dataset $\mathcal{X}_{u_{test}}$, for a total of 502 unlabeled testing profiles.

In our implementation, the batch size for training the supervised discriminator was $N=10$, selected randomly from the labeled dataset each time. Also, 50 randomly selected profiles from the unlabeled dataset with an additional 50 profiles created by the generator network were used to train the  unsupervised discriminator, for a batch size $M=100$. Finally, $P=100$ random genetic profiles were used to train the generator model.

We followed two testing procedures to perform accuracy estimation. The first, denoted by $T_1$, simply measures the accuracy of each of the discriminator outputs, independently. The second, denoted by $T_2$, aggregates both of the discriminator outputs. Testing procedure $T_1$ calculates the labeled accuracy $Acc_{(\mathcal{D}_l)}$ and unlabeled accuracy $Acc_{(\mathcal{D}_u)}$ on the testing data sets $\mathcal{X}_{l_{test}}$ and $\mathcal{X}_{u_{test}}$, respectively. Testing procedure $T_2$ uses the testing dataset $\mathcal{X}_{l_{test}}$ to simulate a deployment scenario. It is a two-step procedure: first, it checks if the discriminator is able to recognize the test profile as a real profile, and calculates the accuracy ${Acc_1}_{(\mathcal{D}_l)}$ on the testing dataset. After that, it verifies if the labeled output matches the actual data label, producing an accuracy estimate ${Acc_2}_{(\mathcal{D}_l)}$ using only the correctly predicted profiles in the first step. The accuracy rates obtained in our experiments according to testing procedures $T_1$ and $T_2$ are displayed in Table~\ref{tab:tests}.

\begin{table}[h!]
\caption{Accuracy rates according to testing procedures $T_1$ and $T_2$.}
\label{tab:tests}
\begin{center}
\begin{tabular}{clcccc}
\hline
Models                                                                &  & \multicolumn{2}{c}{$T_1$} & \multicolumn{2}{c}{$T_2$ } \\
 &  & $Acc_{(\mathcal{D}_l)}$ & $Acc_{(\mathcal{D}_u)}$  & ${Acc_1}_{(\mathcal{D}_l)}$ & ${Acc_2}_{(\mathcal{D}_l)}$ \\[1ex] \hline
${\mathcal{M}_1}$(AFD \textless 0.07) &  & 0.6111       & 0.8719        & 0.9444  & 0.6470 \\
${\mathcal{M}_2}$(AFD \textless 0.10) &  & 0.6999       & 0.9918        & 1.0  & 0.6999 \\
${\mathcal{M}_3}$(AFD \textless 0.21) &  & 0.5263       & 0.9919        & 0.9473  & 0.5555 \\
${\mathcal{M}_4}$(SVM-RFE)            &  & 0.9375       & 0.9430        & 1.0  & 0.9375 \\ \hline
\end{tabular}
\end{center}
\end{table}

The first two columns in Table \ref{tab:tests} from testing procedure $T_1$ show good results for the unlabeled output for all models, meaning that gGAN has good performance when differentiating between real and synthetic data. Models ${\mathcal{M}_2}$, and ${\mathcal{M}_3}$ correctly classified almost all of the test profiles in the $\mathcal{X}_{u_{test}}$ dataset (n=500). Model ${\mathcal{M}_4}$, using the SNPs obtained by the SVM-RFE method, showed superior results, obtaining 94.3\% accuracy when discriminating test profiles in $\mathcal{X}_{u_{test}}$ as real or synthetic, and nearly 93.75\% accuracy on the labeled testing profiles in $\mathcal{X}_{l_{test}}$ (these results are similar to the ones obtained in \cite{davi2019severe}).

The next two columns in Table \ref{tab:tests} from testing procedure $T_2$ show that all models built with AFD SNPs display high accuracy on testing profiles in $\mathcal{X}_{l_{test}}$, which reinforces the AFD hypothesis for selecting the best SNPs. On the other hand, AFD ignores the relationship between the SNPs and the dengue phenotype, as it only considers the similarity between SNPs. This can be noticed in the results for ${Acc_2}_{(\mathcal{D}_l)}$. Model ${\mathcal{M}_4}$ is far more accurate in this respect, which confirms that the SNPs used by ${\mathcal{M}_4}$ have information on the Dengue phenotype.

\section{Conclusion}

In this paper, we introduced a novel approach to train GANs, which is capable of generating labeled genetic datasets using a small labeled dataset and a larger unlabeled dataset, exploiting concepts of semi-supervised learning and data augmentation to create a new approach to deal with the limited labeled data available to researchers. 

The proposed gGAN architecture modifies the usual perspective on GANs: instead of a tool solely designed to generate labeled genetic profiles, it also provides a classifier with a level of confidence. Previous ML methods to classify diseases based on genetic information are not aware of data coming from new populations. With gGAN, one may see the unlabeled output as a self-aware output. While the labeled output of the discriminator predicts whether the individual with the corresponding genetic profile is likely to develop Severe Dengue, the unlabeled output predicts whether the genetic profile is real or synthetic. Given a real genetic profile, if the gGAN classifies it as not real (i.e., synthetic), we can infer that it was most likely never seen before and thus is not appropriate for prediction by the model. 

The differences within the allelic frequencies between populations is a challenge when one is using multiples datasets from different populations. Indeed, even the 1000 Genome Dataset\cite{10002015global} is composed of different populations and is therefore subject to a variety of allelic frequencies itself. Due to limitations on the size of our datasets, we opted to filter the datasets using the AFD, but other approaches would have also been possible if a good sample of both labeled and unlabeled data were available for a population. Since we utilized the labeled dataset from Recife, Brazil \cite{cordeiro2007characterization}, and there is no large unlabeled dataset for this specific population, we had to construct our own AFD thresholds. Another possible approach to this issue would be to use Style Transfer to somehow make the synthetic data with frequencies between those of the two datasets.

Nevertheless, our methods produced good results using the SNPs selected by \cite{davi2019severe} using the same labeled dataset we are using here. In their work, they chose the 13 most representative SNPs to classify Severe Dengue. Indeed, we had good results using those SNPs in the labeled phase of the training and test. However, we also had surprisingly good results using the unlabeled output of the discriminator network, which lead us to believe that the discriminator was able to generalize well, even with the allelic frequencies discrepancies. 

As expected, the results of the model ${\mathcal{M}_4}$ in the tests using $\mathcal{X}_{l_{test}}$ were far better than the models trained using the AFD method to select SNPs. Also, the model ${\mathcal{M}_2}$ had a fair accuracy (almost 70\%) if we think of it as a triage tool for prognosis. Those results are even more acceptable during disease outbreaks, when more specific tests are commonly scarce and expensive. Another important aspect is that it can be applied at any stage of the disease, or even before infection, and can use a broad choice of human sample tissue, since it is only based on genetic profiles.

The COVID-19 pandemic is a good example of a scenario where the method described here could bring great benefits to society. Our approach could be used to help triage patients with a level of confidence. It could also be used to avoid exposure of healthcare professionals who are sensitive to the disease. The proposed method could be a powerful tool to help public health efforts against a new outbreak. 

\bibliographystyle{IEEEtran}
\bibliography{bib}

\end{document}